\documentclass{amsart}
\usepackage{amssymb,amsmath,amsfonts}
\usepackage[foot]{amsaddr}
\usepackage{geometry}
\usepackage{multicol}
\usepackage{graphicx}
\usepackage{enumerate,enumitem}
\usepackage{verbatim,setspace}
\usepackage{pdfsync}
\usepackage{gastex}
\usepackage[lowtilde]{url}
\usepackage[dvipsnames]{xcolor}
\usepackage[utf8]{inputenc}\usepackage[T1]{fontenc}% Polskie literki

\usepackage{syntax} %% <<-- http://texdoc.net/texmf-dist/doc/latex/mdwtools/syntax.pdf : sekcje 1.4, 1.5
\usepackage[colorinlistoftodos]{todonotes}

\begin{document}
\title{Simplified Boardgames}
\author{Jakub Kowalski}
\email{jko@cs.uni.wroc.pl}
\author{Jakub Sutowicz}
\email{jakubsutowicz@gmail.com}
\author{Marek Szyku{\l}a}
\email{msz@cs.uni.wroc.pl}
\address{Institute of Computer Science, University of Wroc{\l}aw, Wroc{\l}aw, Poland}
\begin{abstract}
We formalize Simplified Boardgames language, which describes a subclass of arbitrary board games.
The language structure is based on the regular expressions, which makes the rules easily machine-processable while keeping the rules concise and fairly human-readable.
%\keywords{board, chess, language, game}
\end{abstract}
\maketitle

\section{Introduction}

Simplified Boardgames is a class of fairy-chess-like games, first introduced in~\cite{Bjornsson12Learning}, and slightly extended in~\cite{Kowalski15Testing} (see~\cite{Gregory15TheGRL} for an alternative extension).
The class was developed for the purpose of learning the game rules through the observation of plays. The Simplified Boardgames language describes turn-based, two player, zero-sum games on a rectangular board with piece movements being a subset of a regular language.

Here we provide a formal specification for Simplified Boardgames. Despite the fact that the class has been used in several papers, its formal grammar was still not clearly defined, and some issues were left ambiguous.
Such a definition is crucial for further research concerning AI contests, procedural content generation, translations, etc. For comparison, Metagame system, which can be seen as Simplified Boardgames predecessor, had its grammar explicitly declared in~\cite{Pell92METAGAME}.

\section{Syntax and semantics}

In this section we present the formal grammar for Simplified Boardgames, inspired by the look of training records provided by Bj\"ornsson in his initial work \cite{Bjornsson12Learning}.
The grammar construction is also affected by our experiences concerning Simplified Boardgames, especially in the domain of procedural content generation. The version presented differs only slightly comparing to the versions used in \cite{Kowalski15Testing,Kowalski15Procedural,Kowalski16Evolving}.

\subsection{Grammar}

The formal grammar in EBNF is presented in Figure~\ref{fig:grammar}.
C-like comments can appear anywhere in the game definition: ``//'' starts a line comment and every next character in the line is ignored. ``/*'' starts a multiline comment and every character is ignored until the first occurrence of ``*/''.

\begin{figure}[!ht]
\centering
\begin{grammar}

<sbg> ::= `<\!\,<' <name> `>\!\,>' `<BOARD>' <board> `<PIECES>'  <pieces> `<GOALS>' <goals>

<name> ::= alphanumspace \{alphanumspace\}

<board> ::= <nat> <nat> \{ <row> \}

<row> ::= `|' \{"[.a-zA-Z]"\} `|'

<pieces> ::= \{ "[A-Z]" <regexp> `&' \}

<regexp> ::= <rsc> | <regexp> <regexp> | <regexp> `+' <regexp> | `(' <regexp> `)' [<power>]

<rsc> ::= `(' <int> `,' <int> `,' <on> `)' [<power>]

<power> ::= `^' <nat> | `^' `*'

<on> ::= "[epw]"

<goals> ::= <nat> `&' \{ <goal> `&' \}

<goal> ::= `#' <letter> <nat> | `@' <letter> <squares>

<letter> ::= "[a-zA-Z]"

<squares> ::= <nat> <nat> \{ `,' <nat> <nat> \}

%<statement> ::= <ident> `=' <expr> 
%\alt `for' <ident> `=' <expr> `to' <expr> `do' <statement> 
%\alt `{' <stat-list> `}' 
%\alt <empty> 

%<stat-list> ::= <statement> `;' <stat-list> | <statement> 

\end{grammar}
\caption{Formal grammar for Simplified Boardgames game rules.}
\label{fig:grammar}
\end{figure}

The start non-terminal symbol is ``sbg''.
The ``nat'' non-terminal stands for a natural number (thus a non-empty sequence of digits), while ``int'' stands for a signed integer (thus it is ``nat'' optionally preceded by ``-''). The ``alphanumspace'' non-terminal generates all alphanumerical characters or a space.

\subsection{Example}

An exemplary game called Gardner\footnote{\url{ http://en.wikipedia.org/wiki/Minichess#5.C3.975_chess}}, formatted according to Simplified Boardgames grammar is presented partially in Figure~\ref{fig:gardner}. It is $5\times 5$ chess variant proposed by Martin Gardner in 1969 and weakly solved in 2013 \cite{Mhalla13Gardner} -- the game-theoretic value has been proved to be a draw.

The starting position looks as in the regular chess with removed columns $f$, $g$, $h$, and rows $3$, $4$, $5$. The rules are those of classical chess without the two squares move for pawns, en-passant moves and castling. Additionally, as a countermeasure for not supporting promotions, our implementation provides additional winning condition by reaching the opponent's backrank with a pawn.

\begin{figure}[!ht]
%\begin{minipage}{0.40\textwidth}
 \centering
\begin{verbatim}
<<Simplified Gardner>>  
<BOARD>
5 5
|rnbqk|
|ppppp|
|.....|
|PPPPP|
|RNBQK|
<PIECES>       // P - pawn, R - rook, N - knight, B - bishop, Q - queen, K - king
P (0,1,e) + (-1,1,p) + (1,1,p) &
R (0,1,e)(0,1,e)^* + (0,1,e)^*(0,1,p) + (0,-1,e)(0,-1,e)^* + (0,-1,e)^*(0,-1,p) + 
  (1,0,e)(1,0,e)^* + (1,0,e)^*(1,0,p) + (-1,0,e)(-1,0,e)^* + (-1,0,e)^*(-1,0,p) &
N (2,1,e) + (2,-1,e) + ... + (-1,-2,p) &
B (1,1,e) + (1,1,p) + (1,1,e)^2 + (1,1,e)(1,1,p) + (1,1,e)^3 + (1,1,e)^2(1,1,p) +
  (1,1,e)^4 + (1,1,e)^3(1,1,p) + ... + (-1,-1,e)^4 + (-1,-1,e)^3(-1,-1,p) &
Q (0,1,e)(0,1,e)^* + (0,1,e)^*(0,1,p) + (0,-1,e)(0,-1,e)^* + (0,-1,e)^*(0,-1,p) + 
  (1,0,e)(1,0,e)^* + (1,0,e)^*(1,0,p) + (-1,0,e)(-1,0,e)^* + (-1,0,e)^*(-1,0,p) +
  (1,1,e)(1,1,e)^* + (1,1,e)^*(1,1,p) + (1,-1,e)(1,-1,e)^* + (1,-1,e)^*(1,-1,p) + 
  (1,-1,e)(1,-1,e)^*+(1,-1,e)^*(1,-1,p)+(-1,-1,e)(-1,-1,e)^*+(-1,-1,e)^*(-1,-1,p) &
K (0,1,e) + (0,1,p) + (0,-1,e) + (0,-1,p) + ...  + (-1,-1,e) + (-1,-1,p) &
<GOALS>
100 &
@P 0 4, 1 4, 2 4, 3 4, 4 4 &
@p 0 0, 1 0, 2 0, 3 0, 4 0 &
#K 0 &
#k 0 &
\end{verbatim}
\caption{The Simplified Boardgames version of Gardner.}\label{fig:gardner}
\end{figure}

\subsection{Semantics}

The game is played between two players, \emph{black} and \emph{white}, on a rectangular board. White player is always the first to move.
The board size is given by the two numbers in the \texttt{<BOARD>} section, generated from ``board'' non-terminal, which represents the \emph{width} and the \emph{height}, respectively.
Subsequently the initial position is given: empty squares are represented by dots, white pieces as the uppercase letters, and black pieces as the lowercase letters. To be considered as valid, there must be exactly \emph{height} rows and \emph{width} columns. Although it may be asymmetric, the initial position is given from the perspective of the white player, i.e.\ forward means ``up'' for white, and ``down'' for black.

During a single turn, a player has to make a move using one of his pieces.
Making a move is done by choosing the piece, and change its position according to the specified movement rule for this piece.
At any time, at most one piece can occupy a square, so finishing a move on a square containing a piece (regardless of the owner) results in removing it (capturing). Note that in the situation when the destination square is the starting one, the whole board remains unchanged. No piece addition is possible. After performing a move, the player gives control to the opponent.
%It requires changing the position of a piece accordingly to given movement rules. At any time, at most one piece can occupy a square, so finishing a move on a square containing other piece (regardless of the owner) results in capturing it. No piece addition is possible. After his move, a player gives control to the opponent.

The movement rules of available game pieces are declared in the \texttt{<PIECES>} section and generated from the ``pieces'' non-terminal. One piece can have at most one movement rule, which consists of the letter of the piece and a regular expression.
A piece without the movement rule is allowed but cannot be moved.
For a given piece, the set of legal moves is the set of words described by a regular expression over an alphabet $\Sigma$ containing triplets $(\Delta x, \Delta y, \mathit{on})$, where $\Delta x$ and $\Delta y$ are relative column/row distances, and $\mathit{on} \in\{e, p, w\}$ describes the content of the destination square: $e$ indicates an empty square, $p$ a square occupied by an opponent piece, and $w$ a square occupied by an own piece.
We assume that $x \in \{-\mathit{width}+1,\ldots,\mathit{width}-1\}$, and $y\in \{-\mathit{height}+1,\ldots,\mathit{height}-1\}$, and so $\Sigma$ is finite.

While the piece's owner is defined by the case (upper or lower), its letter encode the piece's type. Pieces with the same type have the same language of legal moves, thus declaration is made for the white pieces only.
Note, however, that a positive $\Delta y$ means forward, which is a subjective direction, and differs in meaning depending on the player. 

%Consider a rule $w \in \Sigma^*$, such that $w=a_1a_2\ldots a_k$, each $a_i=(\Delta x_i, \Delta y_i, \mathit{on}_i)$, and suppose that a piece stands on a square $\langle x, y \rangle$. Then, the rule $w$ is applicable if and only if, for every $i$ such that $1 \le i \le k$, the content condition $\mathit{on}_i$ is fulfilled by the content of the square $\langle x+\sum_{j=1}^i \Delta x_j, y+\sum_{j=1}^i \Delta y_j \rangle$. If move rule $w$ is applicable in the current game position, then the move transferring a piece from $\langle x, y\rangle$ to $\langle x+\sum_{i=1}^k \Delta x_k, y+\sum_{k=1}^k\Delta y_i \rangle$ is legal. 
Consider a piece and a word $w \in \Sigma^*$ that belongs to the language described by the regular expression in the movement rule for this piece.
Let $w=a_1a_2\ldots a_k$, where each $a_i=(\Delta x_i, \Delta y_i, \mathit{on}_i)$, and suppose that the piece stands on a square $\langle x, y \rangle$.
Then, $w$ describes a move of the piece, which is applicable in the current board position if and only if, for every $i$ such that $1 \le i \le k$, the content condition $\mathit{on}_i$ is fulfilled by the content of the square $\langle x+\sum_{j=1}^i \Delta x_j, y+\sum_{j=1}^i \Delta y_j \rangle$.
The move of $w$ changes the position of the piece piece from $\langle x, y\rangle$ to $\langle x+\sum_{i=1}^k \Delta x_k, y+\sum_{k=1}^k\Delta y_i \rangle$.

\begin{figure}[!ht]
\centering
\includegraphics[scale=0.25]{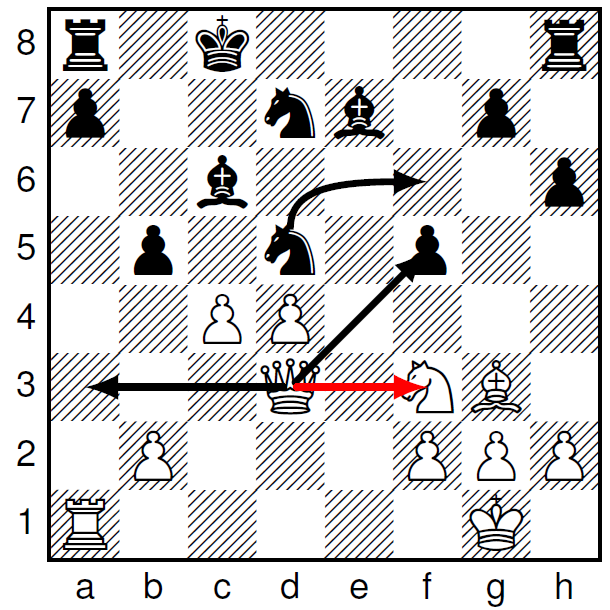}
%\chessboard[setpieces={ra8, kc8, rh8, pa7, nd7, be7, pg7, bc6, ph6, pb5, nd5, pf5, Pc4, Pd4, Qd3, Nf3, Bg3, Pb2, Pf2, Pg2, Ph2, Ra1, Kg1},
%            pgfstyle=knightmove,   markmoves={d5-f6},
%            pgfstyle=straightmove, markmoves={d3-f5, d3-a3},
%            pgfstyle=straightmove, color=red, markmoves={d3-f3}]
\caption{A chess example. Two legal moves for the queen on $d4$ are shown. The capture to $f5$ is codified by a word $(1,1,e)(1,1,p)$, while move to $a3$ is encoded by $(-1, 0,e)(-1,0,e)(-1,0,e)$. The move to $f3$ is illegal, as in the language of queen's moves, no move can end on a square containg own's piece. The $d5-f6$ knight move is a direct jump codified by a one-letter word $(2,1,e)$.}
\label{fig:chessboard}
\end{figure}

In contrast to the Bj\"{o}rnsson's definition, rules where the same square is visited more then once are allowed. Technically, we found this restriction superfluous.
Note that during computation of legal moves, the board position is not changed, so the field with relative coordinates $(0,0)$ always contains the player's moving piece. Hence, $(0, 0, \mathit{w})$ is always legal, while $(0, 0, \mathit{e})$ is always illegal.
An example of how move rules work is shown in Figure~\ref{fig:chessboard}.

%\newchessgame

Lastly, the \texttt{<GOALS>} section provides game terminal conditions.
The first value is the \emph{turnlimit}, whose exceedance automatically causes a draw if no other terminal condition is fulfilled. The turnlimit is given in the so-called ``half-moves'' in chess, i.e.\ the value $t$ means $\lceil\frac{t}{2}\rceil$ moves of the first player and $\lfloor\frac{t}{2}\rfloor$ moves of the second player. A player automatically loses the game when he has no legal moves at the beginning of his turn (e.g.\ because he has no pieces left).

A player can win by moving a certain piece to a fixed set of squares, which are defined by entries denoted by the \texttt{@} symbol. The values are given in absolute coordinates, and $(0, 0)$ square is located in the lower left corner of the board. Alternatively, as introduced in \cite{Kowalski15Testing}, a player can lose if the number of his pieces of a certain type reaches a given amount, defined by entries denoted by the \texttt{\#} symbol. The terminal conditions can be asymmetric.

\section{Summary}

The language can describe many of the fairy chess variants in a concise way. Unlike Metagame, Simplified Boardgames includes games with asymmetry and position-dependent moves (e.g.\ Chess initial double pawn move).
%moves that can capture own pieces. 
The usage of finite automata for describing pieces' rules, and thus to move generation, allows fast and efficient computation of all legal moves given a board setup. The regularity of the description makes it also convenient for e.g.\ procedural content generation \cite{Kowalski15Procedural,Kowalski16Evolving}. However, it causes some important limitations. Actions like castling, en-passant, or promotions are impossible to express, as all the moves depending on the game history. Although absolute position of a piece is not available, this nuisance can be cleverly bypassed, so it is possible to describe e.g.\ chess pawn initial two-square advance.

\bibliographystyle{plain}
\bibliography{ggp}
\end{document}